\pdfoutput=1
\documentclass[11pt]{article}

\usepackage[preprint]{acl}

\usepackage{times}
\usepackage{latexsym}
\usepackage{todonotes}
\usepackage{booktabs}
\usepackage{placeins}
\usepackage{multirow}
\usepackage{graphicx}
\usepackage{svg}

\usepackage[T1]{fontenc}
\usepackage[utf8]{inputenc}

\usepackage{microtype}

\usepackage{inconsolata}

\usepackage{graphicx}

\title{Language Models are Surprisingly Fragile to Drug Names in Biomedical Benchmarks}

\author{
Jack Gallifant$^{1}$\thanks{Co-first authors: Jack Gallifant and Shan Chen}, 
Shan Chen$^{2,3,4}$\footnotemark[1], % Uses the same footnote as Shan Chen
Pedro Moreira$^{1,8}$, Nikolaj Munch$^{1,5}$, Mingye Gao$^{1}$\\ 
\textbf{Jackson Pond$^{2,3}$, Hugo Aerts$^{2,3,7}$, Leo Anthony Celi$^{1,2,9}$,}
\\\textbf{Thomas Hartvigsen$^{1,6}$, Danielle S. Bitterman$^{2,3,4}$\thanks{Corresponding author: dbitterman@bwh.harvard.edu}}\\
\\
$^1$MIT, $^2$Harvard, $^3$Mass General Brigham, $^4$Boston Children's Hospital, \\ $^5$Aarhus University $^6$University of Virginia, $^7$Maastricht University, \\ $^{8}$Universitat Pompeu Fabra, $^{9}$Beth Israel Deaconess Medical Center
}

\begin{document}
\maketitle

% Focus this on medical brand generic as opposed to random synonym replacement 
%% Medical synonyms are context-dependent - therefore, simple synonym replacement is not possible in this context
%% Therefore, expert annotation is a major contribution

\begin{abstract}
Medical knowledge is context-dependent and requires consistent reasoning across various natural language expressions of semantically equivalent phrases. This is particularly crucial for drug names, where patients often use brand names like Advil or Tylenol instead of their generic equivalents. To study this, we create a new robustness dataset, \textbf{RABBITS}, to evaluate performance differences on medical benchmarks after swapping brand and generic drug names using physician expert annotations.

We assess both open-source and API-based LLMs on MedQA and MedMCQA, revealing a consistent performance drop ranging from 1-10\%. Furthermore, we identify a potential source of this fragility as the contamination of test data in widely used pre-training datasets.\footnote{All code is accessible at \url{https://github.com/BittermanLab/RABBITS}, and a HuggingFace leaderboard is available at \url{https://huggingface.co/spaces/AIM-Harvard/rabbits-leaderboard}.}
\end{abstract}

% Introduction

% I suggest 5 paragraphs:
% 1. Motivation: What's the problem and why's it important?
%% Start early to focus on the brand /generic focus
%% LLMs being used in healthcare increasingly both patient facing and in hospitals etc
%% Medical language is inherently synonymous and context specific 
%% implications if performance drops is clear with difference in accuracy of advice and patient harm

% 2. Gap in literature: What have others tried and why have they failed?
% current robustness work has been completed but little focussed on health domain

% 3. Proposed work/problem: What do we do?
% this is particularly due to lack of availability of publically available datates, particularly in healthcare of expert annotated datasets
% we solve this by using existing medical benchmarks
% use the interchangeability of drugs and have experts annotate these

% 4. Findings: What did we find?
%% When performing the swap- average drop was significant XX% 
%% we see they know names with increasing accuracy across model size
% we also see significant amounts of test set data in publically available benchmarks such as dolma 

% 5. Contributions/Implications: What do we add to the literature?
%% New type of robustness evaluation for medical LLMs
%% Dataset that allows for ongoing evaluation
%% Current state of this benchmark

\section{Introduction}
% 1. Motivation: What's the problem and why's it important?
Large Language Models (LLMs) are poised to transform medicine by providing data processing and decision support capabilities \cite{jiang2023health, clusmann2023future}. However, the medical deployment of LLMs demands high accuracy and reliability, as errors can result in severe health consequences \cite{chen2024effect, goodman2024ai, yan2024worse}. A key challenge is the synonymy and context-specific nature of medical language; for instance, patients might use brand names like Advil or Tylenol instead of pharmaceutically equivalent generic terms such as ibuprofen or acetaminophen. LLMs must, therefore, be able to provide consistent and accurate advice in the face of this variability. Fluctuations could lead to risks like medical misinformation, medication errors due to incorrect medication advice, and biases toward or against proprietary products. Our study investigates the effects of substituting drug names—from brand to generic and vice versa—on LLM performance. 

% 2. Gap in literature: What have others tried and why have they failed?
Building on the need for robustness in medical LLM applications, numerous efforts have developed knowledge benchmarks \cite{jin-etal-2019-pubmedqa, hendrycks2021measuring, jin2020disease, liu2023benchmarking, wang2024mmlupro}. Yet, these initiatives primarily tackle general language tasks and often neglect the unique challenges of medical terminology in real-world settings. There is an unmet need to overcome this research gap, as the variability in medical language implies that conventional robustness evaluations might not sufficiently cater to specialized healthcare demands.

% 3. Proposed work/problem: What do we do?
A key reason for this gap is the lack of publicly available, expert-annotated datasets specific to the healthcare domain. To address this issue, our work leverages existing medical benchmarks and employs physician expert annotators to substitute brand names with their generic counterparts and vice versa. 

% 4. Findings: What did we find?
Our findings reveal a surprising drop in the performance of LLMs on common medical benchmarks when the drug names are swapped from generic to brand names: \textbf{4\% drop in accuracy on average}. This is concerning given that patients commonly use brand names and are less likely to spot errors, especially given existing misconceptions that brand drugs are superior to equivalent generics \cite{Colgan2015, Sewell2012}. Furthermore, we identify a potential source for this fragility: Open pretraining datasets contain substantial amounts of benchmark test data.

% 5. Contributions/Implications: What do we add to the literature?
Our research introduces a novel category of robustness evaluation centered on drug name interchangeability. We present \textbf{RABBITS} (\textbf{R}obust \textbf{A}ssessment of \textbf{B}iomedical \textbf{B}enchmarks \textbf{I}nvolving drug \textbf{T}erm \textbf{S}ubstitutions for Language Models) a specialized dataset and leaderboard to aid in evaluating LLM performance in healthcare. Specifically, our study combines and modifies select questions from the MedMCQA \cite{pal2022medmcqa} and MedQA \cite{jin2020disease} benchmarks to:
\begin{itemize}
    \item Assess model robustness in understanding clinical knowledge across drug synonyms.
    \item Detect potential dataset contamination in biomedical benchmarks.
    \item Highlight the importance of robustness to nomenclature variations in the healthcare domain.
\end{itemize}

% Our ultimate aim is to improve the safety and reliability of LLM-driven medical systems in critical healthcare environments.

% Related Work
\section{Related Work}
% Shift 2.2 for closer to medical language model robustness
%% There has been a lot of work on robustness for LLMs more generally 
%% However, there is a shocking paucity of robustness evaluation in the medical domain
%% This is a weakness of the current literature
%% Reason for this is the lack of expert-curated datasets in this medical domain

\subsection{Dataset Contamination}
Dataset contamination in training data is a well-documented issue and can affect the performance and generalizability of LLMs. Many studies have aimed to detect benchmark questions within LLM training data \cite{shi2024detecting, xu2024benchmarking, zhou2023dont}. For instance, research by \citet{recht2019imagenet} illustrated that models trained on contaminated datasets often exhibit inflated performance metrics that do not generalize well to new, unseen data. This problem is particularly concerning for medical LLMs, where inaccurate information can harm patients \cite{chen2024effect, yan2024worse}.

Various strategies have been employed to mitigate dataset contamination. These include removing data with high n-gram overlap with benchmark datasets \cite{brown2020language} and employing embedding similarity to filter out similar data \cite{shi2024detecting}. More advanced approaches involve functional evaluations, such as generating new, unique problem instances for each evaluation \cite{srivastava2024functional}. Addressing contamination is crucial for ensuring that LLMs provide reliable outputs, especially in sensitive domains like healthcare.

\subsection{Evaluating Model Robustness}
LLMs gain broad capabilities from large-scale data ingestion \cite{wei2022emergent}, but this also introduces significant challenges \cite{lu2023emergent, chen2024ocad256}. While larger models often perform better, these improvements are not always consistent across domains \cite{magnusson2023paloma}. Moreover, recent research has questioned the actual reasoning abilities of LLMs, suggesting that their performance may be inflated by dataset contamination rather than genuine problem-solving skills \cite{zhang2024careful}.

Some works have looked into LLMs' robustness in terms of faithfulness \cite{han2024safe} and fairness \cite{zack2024assessing,guevara&chen2024large} under clinical settings. Medfuzz introduced a method to test LLMs' robustness in medical question answering by revealing vulnerabilities through modified benchmark questions \cite{ness2024medfuzz}. However, these studies do not specifically address the unique challenges associated with clinical drug terminology and the relationship between robustness and contamination. Hence, there is a significant gap in evaluating LLM robustness for medical applications, particularly in the context of brand and generic drug name interchangeability. This gap underscores the need for focused robustness evaluations tailored to the healthcare sector.

% Methods
\section{Methodology}

\subsection{Brand-Generic Pairs} 
\begin{figure}[t!]
  \includegraphics[width=\columnwidth]{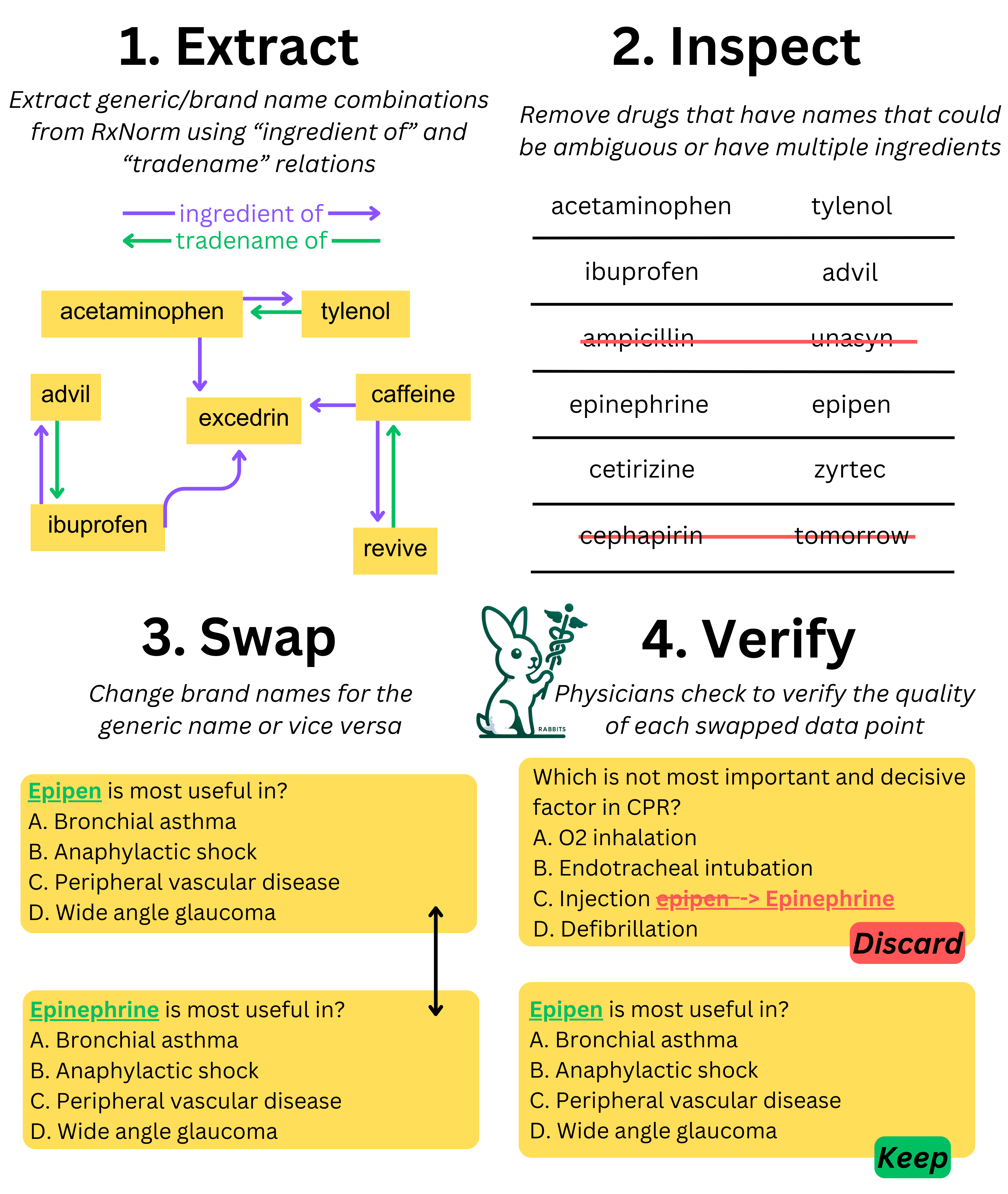}
  \caption{RABBITS dataset generation workflow.}
  \label{fig:workflow}
\end{figure}

Figure \ref{fig:workflow} demonstrates the overall workflow of the study. Appendix \ref{sec:dataset_curation} details the full data quality assurance and dataset curation process. To create the dataset of brand and generic drug name pairs, we used the RxNorm \cite{rxnorm} ontology, which links normalized drug names with many pharmaceutical vocabularies. We extracted combinations of brand and generic drug names using the "ingredient of" and "tradename of" relations, resulting in 2,271 generic drugs mapped to 6,961 brands. For each generic drug, there are often multiple associated brand names. Multiple rounds of expert annotation were performed to derive a final list of 1:1 mapped brand-generic pairs for use in the transformed datasets described below. 

% Two authors (JG, SC) manually examined the dataset to identify and remove any keywords that could overlap with common clinical text. For instance, some brand names, such as "today" or "perform," could lead to erroneous replacements and were thus excluded. Subsequently, the authors (JP, SC) manually reviewed all swapped datasets to ensure their accuracy. Finally, two physician authors (JG, DB) manually verified all swaps to ensure the quality of the final dataset. 

\subsection{Dataset Transformation}
We used regular expressions to identify and replace brand and generic drug names in the questions and answers of MedQA, MedMCQA, MMLU, PubMedQA, and USMLE. MMLU and PubMedQA had fewer than 100 instances of identified drug names in the test split and were excluded from further analysis. USMLE was excluded due to its overlap with MedQA. Thus, the two datasets included in the final RABBITS benchmark are \textbf{MedQA} and \textbf{MedMCQA}. 

The quality of the transformed datasets were iteratively reviewed by 2 physician authors (JG, DB), removing instances where replacements introduced inaccuracies, ambiguities, and/or logical inconsistencies in context. This process is described in detail in Appendix \ref{sec:dataset_curation}. For the rest of the paper, we will refer to the generic-to-brand swapped benchmark as \textbf{g2b} and the brand-to-generic swapped benchmark as \textbf{b2g}. 

To prevent further data contamination, we will not release the full dataset directly. The HuggingFace leaderboard will be the best way to assess new models' robustness in terms of performance. We evaluated the models using the EleutherAI lm-evaluation harness with zero-shot setting \cite{eval-harness}. We forked this repository, added our transformed datasets as new tasks, and made no other modifications. For API models, we used the same prompt format as the lm-evaluation harness with the default hyperparameters. 
%
% All experiments were run using \texttt{transformers} >= 4.37.0, \texttt{cuda} version > 12.0, and \texttt{int4} quantization \cite{dettmers2023qlora}.

Our evaluation focuses on comparing the performance of base models (full list in Appendix Table \ref{tab:models}) across the original and transformed datasets to assess the impact of synonym substitution on accuracy. We report results for g2b due to the limited number of b2g swaps observed. By doing so, we aim to determine whether models can maintain performance despite semantically equivalent pharmaceutical terminology.

All datasets and models used in accordance with owners' licenses.

% Results
%% Counts of drug names
\section{Results and Discussion}
\subsection{Drug Swapping Results}

\begin{figure*}[t!]
  \centering
  \includegraphics[width=0.96\linewidth]{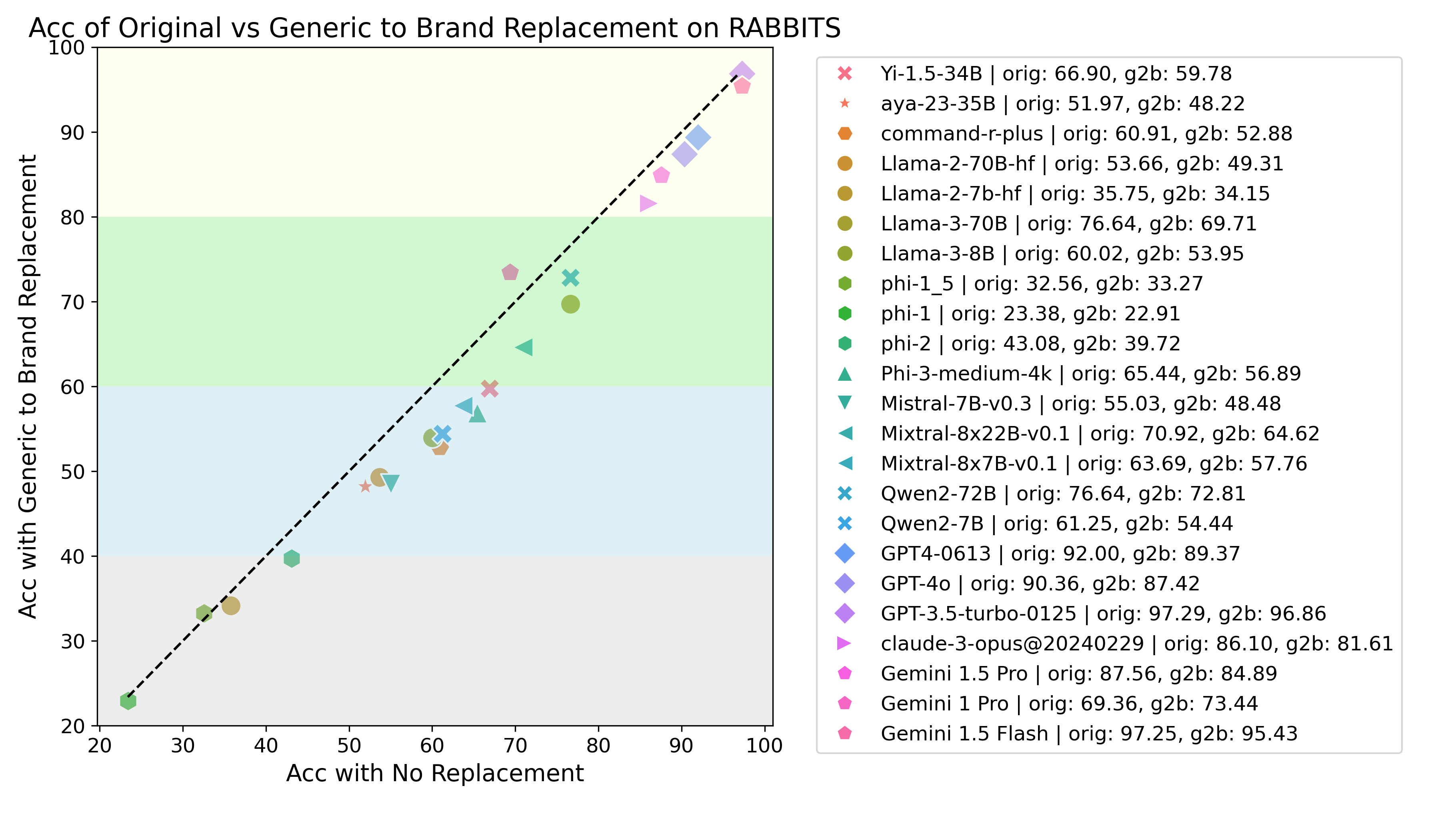}
  \caption{Performance of models on the filtered original datasets compared to the generic-to-brand versions. The dashed diagonal line represents the ideal scenario where synonym swaps do not affect model performance.}
  \label{fig:b4b_g2b_scatter}
\end{figure*}

Figure \ref{fig:b4b_g2b_scatter} presents the performance of each model on the original (no-swap) and transformed (g2b) datasets, alongside the average performance and the difference between the two. The line of robustness, with a gradient of 1, represents the ideal scenario where synonym swaps do not affect the selection of answers. The plot reveals that all open-source models from 7B and above fall below this line, indicating decreased performance when drug names are swapped. We also observe a larger drop among MedMCQA over MedQA across models. Refer to Appendix \ref{sec:appendix_drug_swap} for a detailed breakdown of individual results in Table \ref{tab:all_diffs} and Figure \ref{fig:swap_bar_plot}.

Table \ref{tab:all_diffs} shows that most models experience a decrease in accuracy when generic names are swapped with brand names across different datasets and model sizes. Among large open-source models, the Llama-3-70B model, despite being one of the larger and more accurate models on the original dataset (no-swap accuracy of 76.6\%), decreases to 69.7\% accuracy with generic-to-brand swaps. Overall, API models perform better than their open-source counterparts with higher accuracy and lower performance drop. While larger open-source series like Qwen2, Llama, and Mixtral are more accurate on original datasets, they exhibit greater sensitivity to g2b swaps. This suggests limitations in true comprehension and reasoning abilities.

%% Drug knowledge

\subsection{Model Knowledge of Drug Pairs via Multi-Choice Questions}

We evaluate whether models are able to directly map brand-to-generic drug pairs and vice versa using multiple-choice questions for all drugs that were swapped in our final benchmark dataset. Overall, a clear "scaling law" \cite{kaplan2020scaling} is observed in Appendix \ref{sec:appendix_b4bqa} Figure \ref{fig:b4bqa}, where larger models (active parameter size over 13B) consistently outperform smaller models on this task, with larger open-source and API models achieving accuracy over 97\%. 

%% Drug swapping results
\subsection{Generic and Brand Mentions in Benchmarks and Pre-training Datasets}

Table \ref{tab:dataset_counts} shows our overall dataset swapping statistics where we observe benchmark questions overwhelmingly use generic terms. We also use Infini-gram \cite{Liu2024InfiniGram} to screen the common open-sourced pre-training data, including Redpajama \cite{together2023redpajama}, C4 train \cite{2019t5}, Pile train \cite{gao2020pile}, and Dolma 1.6 \cite{dolma} for drugs identified in RxNorm, filtered for terms that overlap with common terms (Appendix \ref{sec:dataset_curation}, Step 1). Generic names are more common than brand names in these pre-training datasets, as Appendix \ref{sec:screening1} table \ref{tab:infi_gram_filtered} shows. 

\begin{table}[htbp]
    \centering
    \caption{Original and Swapped Dataset Statistics}
    \label{tab:dataset_counts}
    \begin{tabular}{lcccc}
        \toprule
        \multirow{2.5}{*}{\textbf{Dataset}} & \multirow{2.5}{*}{\textbf{Orig.}} & \multirow{2.5}{*}{\textbf{Kept}} & \multicolumn{2}{c}{\textbf{Swap direction}} \\
        \cmidrule(lr){4-5}
         & & & b2g & g2b \\
        \midrule
        MedQA & 1,271 & 378 & 5 & 816 \\
        MedMCQA & 4,180 & 348 & 24 & 626 \\
        % MMLU & 308 & 56 & 0 & 54 \\
        \bottomrule
    \end{tabular}
\end{table}

%% Bio Tuned models results

\subsection{Contamination Source from Pre-training Dataset}
% The top 3 accessible models on the leaderboard (all higher than GPT-4 and Med-Palm on June 1st, 2024) \cite{openlifescienceai/open_medical_llm_leaderboard}, are Llama3 fine-tuned variants.  The Llama3-8B-sft2 model achieved the highest original score of 0.85 but showed a performance drop of -0.15 in the g2b category, similar to Llama3-8B-sft1. Both performance decreases are greater than those observed in the base model version. Similar patterns were seen in the Llama-80B models.

To investigate why we see larger performance drops in MedMCQA than MedQA, we use Infini-gram API \cite{Liu2024InfiniGram} to identify overlaps with the Dolma 1.6 dataset (3.1T tokens) using size 8 n-grams. Each question's n-grams are generated and queried through the Infini-gram API. 

Dataset contamination are 99.21\% and 34.13\% in the MedQA and MedMCQA test datasets, respectively, as Table \ref{tab:infi_gram_contem1} shows. We also benchmark OLMo-1.7-7B-hf, trained only on Dolma, which shows no drop in MedQA (31.22) scores compared to a 3\% drop in MedMCQA (40.90 to 37.93). This likely explains the greater drop in performance in MedMCQA rather than MedQA across models (Appendix \ref{sec:appendix_drug_swap} Figure \ref{fig:swap_bar_plot}).

% \begin{table}[ht]
% \centering
% \caption{Counts and Percentage of contamination of Dolma  dataset with questions from different splits of multiple datasets}
% \begin{tabular}{lrrrr}
% \toprule
% \textbf{Dataset} & \textbf{Split} & \textbf{Total} & \textbf{Contaminated} & \textbf{Percentage}  \\
% \midrule
% \multirow{2}{*}{MedQA} & Train & 4,014 & 3,489 & 86.92\% \\
% & Val & 474 & 465 & 98.10\% \\
% & Test & 508 & 504 & 99.21\% \\
% \midrule
% \multirow{2}{*}{MedMCQA} & Train & 27,766 & 6,222 & 22.41\% \\
% & Val & 504 & 172 & 34.13\% \\
% & Test & 780 & 91 & 11.67\% \\
% \bottomrule
% \end{tabular}
% \label{tab:infi_gram_contem1}
% \end{table}

\begin{table}[ht]
\centering
\caption{Percentage of contamination of MedQA and MedMCQA benchmarks in Dolma dataset}
\begin{tabular}{lrr}
\toprule
\textbf{Dataset} & \textbf{Percentage} \\
\midrule
MedQA Train & 86.92\% \\
MedQA Val & 98.10\% \\
MedQA Test  & 99.21\% \\
\midrule
MedMCQA Train & 22.41\% \\
MedMCQA Val/Test & 34.13\% \\
% MedMCQA Test & 780 & 11.67\% \\
\bottomrule
\end{tabular}
\label{tab:infi_gram_contem1}
\end{table}

% When synonyms—such as swapping generic drug names with their brand counterparts—are introduced, the models are forced to rely on their understanding of context rather than memorization. However, due to the overwhelming presence of original training data within the test set, the models' ability to generalize beyond memorized responses is severely compromised. This is evident from the observed performance drops; the models struggle to maintain accuracy when confronted with even slight variations in terminology introduced through synonym swaps. 

% INPUT TABLE HERE %

\FloatBarrier  % This ensures that the table is placed here

% Conclusion
\section{Conclusion}
We find decreased performance on common medical benchmarks when using different names for the same drug, despite LLMs' ability to match these names, and that these trends scale with LLM size. This suggests that LLM performance may be driven by memorization and not reasoning ability. RABBITS underscores the importance of dataset contamination and model robustness evaluations, particularly in the medical domain. Future research should refine strategies and explore new methods for robustness and fairness evaluation.

% Acknowledgements
\newpage
\section{Limitations}
Our evaluation is limited to biomedical datasets and focuses only on pharmaceuticals. Future work will extend this approach to other medical synonyms. Although the dataset is smaller, trained physicians have curated it multiple times, ensuring its validity and the accuracy of questions after replacement. 
Among the pre-training dataset contamination section, we acknowledge none of these models are trained specifically among the pile, C4, RedPajama, or Dolma. However, we use this as a reasonable proxy for estimating the internet distribution.

\section*{Acknowledgments}
The authors also acknowledge financial support from the Woods Foundation (DB, SC, HA) NIH (NIH-USA U54CA274516-01A1 (SC, HA, DB), NIH-USA U24CA194354 (HA), NIH-USA U01CA190234 (HA), NIH-USA U01CA209414 (HA), NIH-USA R35CA22052 (HA), NIH-USA U54 TW012043-01 (JG, LAC), NIH-USA OT2OD032701 (JG, LAC), NIH-USA R01EB017205 (LAC), DS-I Africa U54 TW012043-01 (LAC), Bridge2AI OT2OD032701 (LAC), NSF ITEST 2148451 (LAC) and the European Union - European Research Council (HA: 866504)

The authors also thank the Google Gemma research grant for supporting the evaluation of the Claude3 and Gemini series models.

% Bibliography entries for custom only
\bibliography{custom}

% Appendix
\onecolumn
\appendix

%%%%%%%% SECTION %%%%%%%%
\section{Brand-Generic Pair Generation and Transformed Dataset Curation}
\label{sec:dataset_curation}
To create the initial dataset of brand and generic drug name pairs, we used the RxNorm \cite{rxnorm} ontology, which links normalized drug names with many pharmaceutical vocabularies. We extracted combinations of brand and generic drug names using the "ingredient of" and "tradename of" relations, resulting in 2,271 unique generic names mapped to 6,961 brands. For each generic name, there are often multiple brand names. Keywords were identified using regular expressions and counted in each question within each column (questions and answer choices), and within each dataset split. Datasets with fewer than 100 instances of identified keywords in the test set were excluded from further analysis. We then carried our the following steps to arrive at our dataset of matched brand and generic drug names, and our final transformed QA datasets, as described below.

\begin{enumerate}
    \item Two authors (SC and JG) reviewed the brand and generic names retrieved from RxNorm and removed keywords that could overlap with common text not referring to drugs. For example, some brand names such as "today" and "perform" could lead to erroneous replacement and were excluded. This resulted in 581 generic keywords mapped to 4297 unique brand keywords. 
    \item Regular expressions were used to to identify and replace names in the medical QA datasets to create 2 initial transformed datasets: generic-to-brand swapped in context, and brand-to-generic swapped in context. 
    \item Two authors (SC and JP) reviewed the resulting datasets to ensure that the replacements were done correctly.
    \item Two physician authors (JG and DB) reviewed the datasets, and identified several areas of ambiguity, errors, and inconsistencies resulting from the regular expression replacement, most commonly: (1) brand names for combination medications swapped to generic names referring to single-agent medications; (2) generic names of combination medications in which a single-agent brand name was swapped for one or both of the drugs in the generic description, for example trimethoprim/sulfamethoxazole replaced with proloprim/sulfamethoxazole; (3) brand names for veterinary formulations; (4) brand names for specific formulations that did not make logical or clinical sense in context, such as brand names for topical formulations replacing descriptions of the drug administered intravenously (either explicitly or implicitly given the clinical context); (5) drug names that are also naturally occurring physiologic compounds, such as amino acids (e.g., tyrosine), vitamins (e.g., Vitamin A); endogenous hormones (e.g., insulin, thyroxine), essential elements (e.g., copper, calcium), etc; and (6) drug names that are also dietary compounds (e.g., caffeine, tryptophan). These questions were annotated, and the drug in question tracked.
    \item Given the above identified errors, to facilitate expert review the drug pairs from Step 1 were provided to GPT-4o, which was prompted to check if the brand name's main component is the paired generic drug, and if the brand name drug is used mainly used for humans. 1247 brand drugs were filtered out of the keyword list. This resulted in 563 generic drug mapped to 3050 brand keywords.
    \item The remaining drugs in Step 5 were provided to Cohere-RAG, which was prompted to provided a list of brand names for each generic name.
    \item A physician author (JG) reviewed the retrieved brand names and selected a single brand name to pair with each generic name. This resulted in 525 generic-to-brand pairs.
    \item These 525 generic-to-brand pairs were used to regenerate the transformed dataset using regular expressions to identify and replace names.
    \item Two physician authors (JG and DB) reviewed the dataset generated in Step 8, and removed any remaining questions where the replacement resulted in ambiguity or inconsistencies in context to ensure quality of the final dataset. 
\end{enumerate}

%%%%%%%% SECTION %%%%%%%%
\newpage
\section{Drug Knowledge}
\label{sec:appendix_b4bqa}

\begin{figure*}[htbp]
  \centering
  \includegraphics[width=0.96\linewidth]{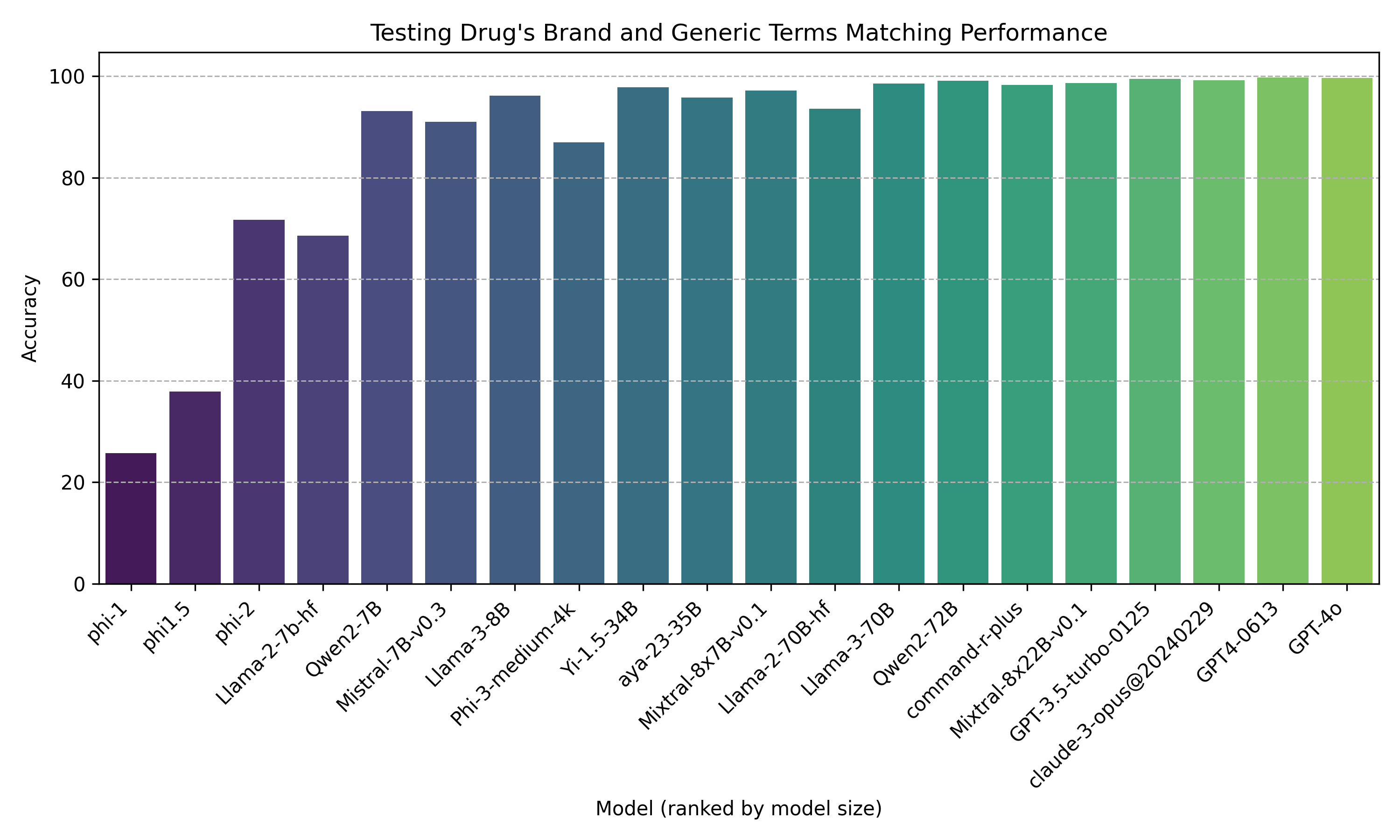}
  \caption{Performance of models on multi-choice question identification of brand-generic drug pairs ordered in increasing model size.  Gemini results are missing due to Google's API safety filters.}
  \label{fig:b4bqa}
\end{figure*}

%%%%%%%% SECTION %%%%%%%%
\newpage
\section{Drug Swapping results}
\label{sec:appendix_drug_swap}

\begin{figure*}[htbp]
  \centering
  \includegraphics[width=0.96\linewidth]{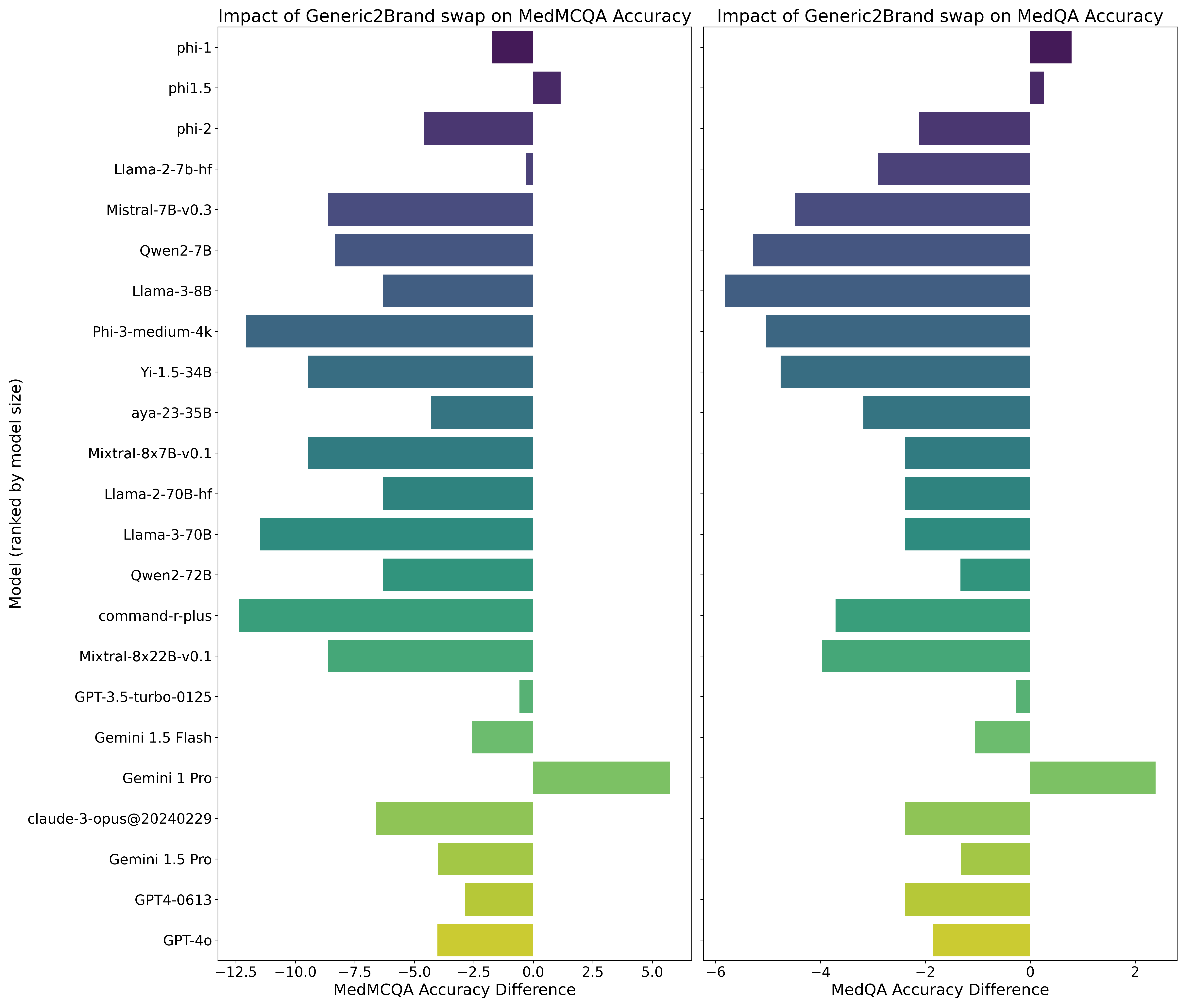}
  \caption{Performance of models on the filtered original datasets compared to the generic-to-brand versions for MedMCQA and MedQA subsets. Negative values indicate worse performance on the swapped dataset.}
  \label{fig:swap_bar_plot}
\end{figure*}

Figure \ref{fig:swap_bar_plot} shows the performance change from the original filtered datasets compared to the dataset where generic drug names are swapped with the brand names (g2b). Notice that the MedMCQA (left) x-axis range is much larger than the MedQA x-axis range, indicating a larger drop across models.

\newpage

\begin{table}[ht!]
\begin{center}
\caption{List of Models Used in Experiments, June 12th, 2024}
\label{tab:models}
\begin{tabular}{lccc}
\toprule
Model Name & Size & MoE & Multi-Modal\\
\midrule
phi-1 \cite{gunasekar2023textbooks} & 1.3B & No & No\\
phi-1.5 \cite{textbooks2} & 1.3B & No & No\\
phi-2 \cite{textbooks2} & 2.7B & No & No\\
phi-3-medium \cite{abdin2024phi} & 14B & No & No\\
Llama3-8B \cite{llama3modelcard} & 8B & No & No\\
Llama3-70B \cite{llama3modelcard} & 70B & No & No\\
llama-2-70B \cite{touvron2023llama} & 70B & No & No\\
llama-2-7B \cite{touvron2023llama} & 7B & No & No\\
c4ai-aya-23-35B \cite{aryabumi2024aya} & 35B & No & No\\
c4ai-r-plus & 104B & No & No\\
Mistral-7B-v0.3 \cite{jiang2023mistral} & 7B & No & No\\
mixtral-8x22B \cite{jiang2024mixtral} & 176B & Yes & No\\
mixtral-8x7B \cite{jiang2024mixtral} & 56B & Yes & No\\
qwen2-72B \cite{qwen2} & 72B & No & No\\
qwen2-7B \cite{qwen2} & 7B & No & No\\
yi-1.5-34B \cite{young2024yi} & 34B & No & No\\
GPT-3.5-turbo-0125 & NA & NA & No\\
GPT4-0613& NA & NA & Yes\\
GPT-4o & NA & NA & Yes\\
Claude 3 Opus & NA & NA & Yes\\
Gemini 1 Pro  & NA & NA & Yes\\
Gemini 1.5 Flash & NA & NA & Yes\\
Gemini 1.5 Pro & NA & NA & No\\
\bottomrule
\end{tabular}
\end{center}
\end{table}
% Table \ref{tab:models} shows a list of all the models that were tested on the RABBITS framework.

\FloatBarrier
% Table of all models we use
\begin{table}[ht]
\centering
\caption{Overall and difference of Model performance on RABBITS}
\label{tab:all_results}
\begin{tabular}{lrrrr}
\toprule
Dataset & Model & g2b & original \\
\midrule
medmcqa & GPT-3.5-turbo-0125 & 97.70 & 98.28 \\
medmcqa & GPT-4o & 86.49 & 90.52 \\
medmcqa & GPT4-0613 & 88.79 & 91.67 \\
medmcqa & Gemini 1 Pro  & 73.85 & 68.10 \\
medmcqa & Gemini 1.5 Flash & 94.83 & 97.41 \\
medmcqa & Gemini 1.5 Pro & 82.47 & 86.49 \\
medmcqa & Llama-2-70B-hf & 45.98 & 52.30 \\
medmcqa & Llama-2-7b-hf & 33.91 & 34.20 \\
medmcqa & Llama-3-70B & 66.67 & 78.16 \\
medmcqa & Llama-3-8B & 52.87 & 59.20 \\
medmcqa & Mistral-7B-v0.3 & 48.28 & 56.90 \\
medmcqa & Mixtral-8x22B-v0.1 & 61.78 & 70.40 \\
medmcqa & Mixtral-8x7B-v0.1 & 55.46 & 64.94 \\
medmcqa & Phi-3-medium-4k & 60.34 & 72.41 \\
medmcqa & Qwen2-72B & 71.55 & 77.87 \\
medmcqa & Qwen2-7B & 55.17 & 63.51 \\
medmcqa & Yi-1.5-34B & 59.77 & 69.25 \\
medmcqa & aya-23-35B & 48.56 & 52.87 \\
medmcqa & claude-3-opus@20240229 & 79.89 & 86.49 \\
medmcqa & command-r-plus & 49.14 & 61.49 \\
medmcqa & phi-1 & 24.14 & 25.86 \\
medmcqa & phi-2 & 37.64 & 42.24 \\
medmcqa & phi1.5 & 31.61 & 30.46 \\
medqa 4options & GPT-3.5-turbo-0125 & 96.03 & 96.30 \\
medqa 4options & GPT-4o & 88.36 & 90.21 \\
medqa 4options & GPT4-0613 & 89.95 & 92.33 \\
medqa 4options & Gemini 1 Pro  & 73.02 & 70.63 \\
medqa 4options & Gemini 1.5 Flash & 96.03 & 97.09 \\
medqa 4options & Gemini 1.5 Pro & 87.30 & 88.62 \\
medqa 4options & Llama-2-70B-hf & 52.65 & 55.03 \\
medqa 4options & Llama-2-7b-hf & 34.39 & 37.30 \\
medqa 4options & Llama-3-70B & 72.75 & 75.13 \\
medqa 4options & Llama-3-8B & 55.03 & 60.85 \\
medqa 4options & Mistral-7B-v0.3 & 48.68 & 53.17 \\
medqa 4options & Mixtral-8x22B-v0.1 & 67.46 & 71.43 \\
medqa 4options & Mixtral-8x7B-v0.1 & 60.05 & 62.43 \\
medqa 4options & Phi-3-medium-4k & 53.44 & 58.47 \\
medqa 4options & Qwen2-72B & 74.07 & 75.40 \\
medqa 4options & Qwen2-7B & 53.70 & 58.99 \\
medqa 4options & Yi-1.5-34B & 59.79 & 64.55 \\
medqa 4options & aya-23-35B & 47.88 & 51.06 \\
medqa 4options & claude-3-opus@20240229 & 83.33 & 85.71 \\
medqa 4options & command-r-plus & 56.61 & 60.32 \\
medqa 4options & phi-1 & 21.69 & 20.90 \\
medqa 4options & phi-2 & 41.80 & 43.92 \\
medqa 4options & phi1.5 & 34.92 & 34.66 \\
\bottomrule
\end{tabular}
\end{table}

\clearpage

\begin{table}[ht!]
\centering
\caption{Overall and difference of Model performance on RABBITS}
\label{tab:all_diffs}
\begin{tabular}{lrrrr}
\toprule
Model & Original & g2b & Average & Difference \\
\midrule
GPT-3.5-turbo-0125 & 97.29 & 96.86 & 97.08 & -0.42 \\
GPT-4o & 90.36 & 87.42 & 88.89 & -2.94 \\
GPT4-0613 & 92.00 & 89.37 & 90.69 & -2.63 \\
Gemini 1 Pro  & 69.36 & 73.44 & 71.40 & 4.07 \\
Gemini 1.5 Flash & 97.25 & 95.43 & 96.34 & -1.82 \\
Gemini 1.5 Pro & 87.56 & 84.89 & 86.22 & -2.67 \\
Llama-2-70B-hf & 53.66 & 49.31 & 51.49 & -4.35 \\
Llama-2-7b-hf & 35.75 & 34.15 & 34.95 & -1.60 \\
Llama-3-70B & 76.64 & 69.71 & 73.18 & -6.93 \\
Llama-3-8B & 60.02 & 53.95 & 56.99 & -6.08 \\
Mistral-7B-v0.3 & 55.03 & 48.48 & 51.76 & -6.55 \\
Mixtral-8x22B-v0.1 & 70.92 & 64.62 & 67.77 & -6.29 \\
Mixtral-8x7B-v0.1 & 63.69 & 57.76 & 60.72 & -5.93 \\
Phi-3-medium-4k & 65.44 & 56.89 & 61.16 & -8.55 \\
Qwen2-72B & 76.64 & 72.81 & 74.72 & -3.83 \\
Qwen2-7B & 61.25 & 54.44 & 57.84 & -6.82 \\
Yi-1.5-34B & 66.90 & 59.78 & 63.34 & -7.12 \\
aya-23-35B & 51.97 & 48.22 & 50.09 & -3.75 \\
claude-3-opus@20240229 & 86.10 & 81.61 & 83.85 & -4.49 \\
command-r-plus & 60.91 & 52.88 & 56.89 & -8.03 \\
phi-1 & 23.38 & 22.91 & 23.15 & -0.47 \\
phi-2 & 43.08 & 39.72 & 41.40 & -3.36 \\
phi1.5 & 32.56 & 33.27 & 32.91 & 0.70 \\
\bottomrule
\end{tabular}
\end{table}
% Table \ref{tab:all_diffs} shows the raw scores for the generic to brand and original filtered datasets for each model.

\FloatBarrier

%%%%%%%% SECTION %%%%%%%%
% \newpage
\clearpage
\section{Pre-training data screening}
\label{sec:screening1}
% Table stats for all infini-grams 
\begin{table}[ht]
\centering
\caption{Statistics for occurrence counts of selected brand and generic terms among popular pre-training datasets}
\begin{tabular}{lrrrr}
\toprule
\textbf{Subset} & \textbf{Terms} & \textbf{Average} & \textbf{Median} & \textbf{Std. Dev} \\
\midrule
\multirow{2}{*}{Dolma} & Generic & 564,151 & 136,682 & 2,399,928 \\
 & Brand & 234,138 & 698 & 2,543,075 \\
\midrule
\multirow{2}{*}{Red Pajama} & Generic & 161,227 & 42,549 & 620,393 \\
 & Brand & 29,561 & 84 & 232,661 \\
\midrule
\multirow{2}{*}{Pile train} & Generic & 96,309 & 28,074 & 325,307 \\
 & Brand & 4,757 & 19 & 43,613 \\
\midrule
\multirow{2}{*}{C4 train} & Generic & 27,973 & 5,454 & 144,162 \\
 & Brand & 9,941 & 26 & 96,504 \\
\bottomrule
\end{tabular}
\label{tab:infi_gram_filtered}
\end{table}

% \input{tables/infi_gram_all}

%%%%%%%% SECTION %%%%%%%%
\newpage
\section{Biomedical-Supervised Fine-Tuned (SFT) Models}

The top 3 accessible models on the leaderboard (all higher than GPT-4 and Med-Palm as of June 1st, 2024) \cite{openlifescienceai/open_medical_llm_leaderboard} are all Llama3 SFT variants. The Llama3-8B-sft2 model achieved the highest original score of 0.85 but showed a performance drop of -0.15 in the g2b category, similar to Llama3-8B-sft1. Both performance decreases are greater than those observed in the base model version. Similar patterns were seen in the Llama-70B models. These degradations among benchmark datasets may be helpful in inspecting SFT models that are over-fitted. 

\label{sec:screening2}
% Table stats for all infini-grams 

\begin{table}[ht]
    \centering
    \caption{Llama-3 Vanilla v.s its Fine-Tuned Variants}
    \label{tab:sft_comparison}
    \begin{tabular}{lccc}
        \toprule
        Model & None $\uparrow$ & g2b $\uparrow$ & $\delta\downarrow$ \\
        \midrule
        llama-3-8B (vanilla) & 0.60 & 0.53 & -0.07 \\
        llama-3-8B-sft1 & 0.80 & 0.66 & -0.14$\uparrow$ \\
        llama-3-8B-sft2 & 0.85 & 0.70 & -0.15$\uparrow$ \\
        \midrule
        llama-3-70B (vanilla) & 0.77 & 0.70 & -0.07 \\
        llama-3-70B-sft1 & 0.75 & 0.66 & -0.09$\uparrow$ \\
        \bottomrule
    \end{tabular}
\end{table}

\end{document}